\documentclass{bmvc2k}
\usepackage{nicefrac,xfrac}

\usepackage{mathtools}
\DeclarePairedDelimiter{\norm}{\lVert}{\rVert}
\usepackage{multirow}
\usepackage{graphicx}
\usepackage{soul}
\usepackage[inline]{enumitem}
\usepackage{siunitx}
\usepackage{microtype}
\usepackage{graphicx}
\usepackage{subfigure}
\usepackage{booktabs} % for professional tables

\usepackage{hyperref}
\usepackage{amsmath}
\usepackage{amssymb}
\usepackage{mathtools}
\usepackage{amsthm}

\title{Improving Human Motion Plausibility \\ with Body Momentum}
\addauthor{Ha Linh Nguyen}{hlinhn@comp.nus.edu.sg}{1}
\addauthor{Tze Ho Elden Tse}{eldentse@comp.nus.edu.sg}{1}
\addauthor{Angela Yao}{ayao@comp.nus.edu.sg}{1}
\addinstitution{
National University of Singapore, \\
Singapore
}
\runninghead{Nguyen et al}{Body Momentum in Human Motion}

\begin{document}
\maketitle

\begin{abstract}

Many studies decompose human motion into local motion in a frame attached to the root joint and global motion of the root joint in the world frame, treating them separately. However, these two components are not independent. Global movement arises from interactions with the environment, which are, in turn, driven by changes in the body configuration. Motion models often fail to precisely capture this physical coupling between local and global dynamics, while deriving global trajectories from joint torques and external forces is computationally expensive and complex. 
To address these challenges, we propose using whole-body linear and angular momentum as a constraint to link local motion with global movement. Since momentum reflects the aggregate effect of joint-level dynamics on the body's movement through space, it provides a physically grounded way to relate local joint behavior to global displacement. 
Building on this insight, we introduce a new loss term that enforces consistency between the generated momentum profiles and those observed in ground-truth data.
Incorporating our loss reduces foot sliding and jitter, improves balance, and preserves the accuracy of the recovered motion. Code and data are available at the \href{https://hlinhn.github.io/momentum_bmvc/}{project page}.
\end{abstract}
\section{Introduction}
\label{sec:intro}

Accurately modeling global human motion is crucial for predicting and analyzing human activities, enabling advancements in applications such as robotics, virtual and augmented reality, and autonomous systems. This process requires not only realistic combinations of joint rotations - commonly emphasized in motion recovery and prediction tasks \cite{Muhammad2022ARO} - but also a plausible trajectory of the root joint. These two components are inherently connected: the center of mass (CoM) of the body moves in response to environmental interactions, which are influenced by changes in joint configurations. 
Plausible motions
must maintain consistency between global trajectories and local joint movements. For instance, a sideways stepping motion should correspond to a lateral global trajectory rather than one indicative of forward or backward walking. Similarly, in a high jump or a back flip, due to the conservation of angular momentum, drawing one’s limbs closer while spinning should increase the global rotation speed, not decrease it.
However, many works treat these two components of human motion separately~\cite{he2022nemf,Li2021TaskGenericHH} or leave their relationship to be inferred from data alone~\cite{Rempe2021HuMoR3H,Kocabas2019VIBEVI}, leading to incompatibilities that render the motion implausible.  A common manifestation of mismatches between the global root trajectory and joint configurations is unexplained root movements. For instance, the root may translate without corresponding changes in contact states, causing unnatural sliding of the feet or other grounded body parts. 
\begin{figure}[ht]
\centering
   \includegraphics[width=\linewidth]
                   {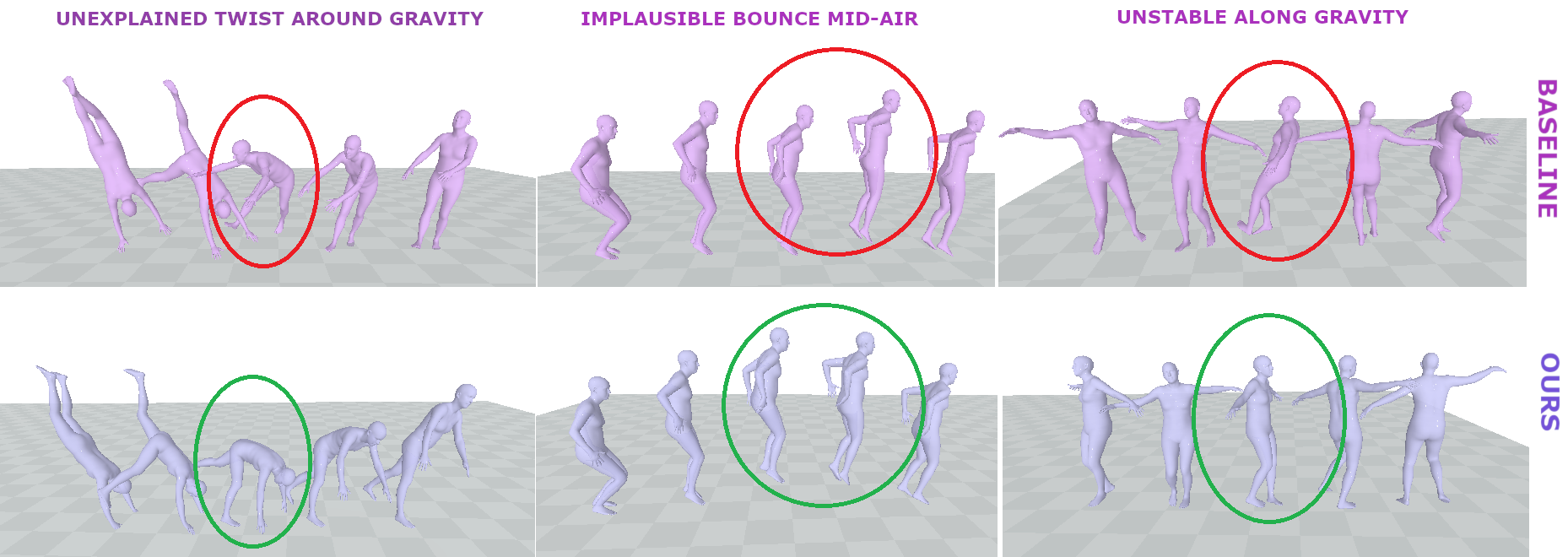}
\vskip -0.15in
\caption{Root movements unexplained by local joint configurations lead to implausible motion (left to right: WHAM \cite{Shin2023WHAMRW}, PhysPT \cite{Zhang2024PhysPTPP}, GLAMR \cite{Yuan2021GLAMRGO}). Our loss encourages more natural motion.}
\label{fig:base-vs-ours}

\end{figure}

To address this issue, three main strategies have emerged: physics-based methods, learning-based approaches, and enhanced kinematic models.
Physics-based methods use simulators to model human motion dynamics~\cite{Yuan2021SimPoESC,Shimada2020PhysCap,Shimada2021NeuralM3}, but they rely on detailed environmental knowledge to compute interaction forces and timing. This often requires simplified representations of the body or surroundings~\cite{Huang2022NeuralMN,Shimada2020PhysCap}, limiting their effectiveness in complex, dynamic, or cluttered scenes. Moreover, optimizing contact forces and timing remains difficult due to their discrete nature.
Learning-based methods attempt to predict these dynamic quantities or ease the optimization process~\cite{Xie2021PhysicsbasedHM,li2022dnd,Zhang2024PhysPTPP,Cong2023EfficientHM}, yet they struggle with limited ground truth data. Annotations generated via inverse dynamics are often unreliable, as they rely on residual forces to correct modeling inaccuracies~\cite{Yuan2020ResidualFC,Werling2024AddBiomechanicsDC}.
Alternatively, some approaches improve the realism of purely kinematic motion by targeting specific elements crucial for plausibility - such as 
balance~\cite{tripathi2024humos}, ground interaction pattern~\cite{Ma2023GraMMaRGM}, contact or friction~\cite{Zhang2021LearningMP}. 
While effective at reducing artifacts like foot sliding or interpenetration, they tend to focus on body-ground interactions and may overlook dynamics in aerial or acrobatic motions like those in basketball, parkour, or gymnastics.

During intervals between contact events, 
the body's rotation is entirely determined by the joint configurations, which influence its moment of inertia. Meanwhile, the translation of the CoM remains unaffected by changes in joint arrangements, as linear momentum is conserved. Even in more common cases with frequent ground contact where momentum is generally not conserved, there are discernible patterns in how momentum terms evolve over time~\cite{Herr2008AngularMI,Negishi2023RegulationOW,Reisman2002CoordinationUT,Chiovetto2018LowdimensionalOO,Maldonado2018OnTC}. Specifically, the angular momentum about the CoM is regulated to be close to $0$ in a large class of movements~\cite{Popovic2004AngularMR,Negishi2023RegulationOW}, while the control of linear momentum implies the control of the CoM trajectory, which tends to remain inside or close to the base of support for balance \cite{Macchietto2009MomentumCF}. %\LN{aim to keep the centroidal momentum stable}.
These patterns establish a relationship between joint movements as observed in the root frame, and the global motion of the body.
From these observations and insights, we propose a novel loss term that provides an explicit connection between global movement to local joint configuration. 
Unlike prior physics-based methods that emphasize modeling forces and torques—which are often unobservable or difficult to estimate from real-world motion data—our approach leverages more accessible, yet still physically meaningful, features of human motion. 
Specifically, we guide the generation or reconstruction of motion by encouraging alignment with ground truth data in terms of whole-body linear and angular momentum, computed in a fixed world frame. These two global motion features capture the overall translation and rotation of the body in a way that is tightly coupled with local joint behaviors, and are critical for producing physically plausible and visually natural motion. By incorporating this loss into our training objective, we ensure that the resulting motions are not only kinematically valid but also globally consistent with how real bodies move in space.
Integrating our loss into existing methods leads to more plausible motion, as evidenced by reduced foot sliding, less unwanted jitter, and greater body stability, all without compromising performance metrics such as accuracy.

To summarize: \begin{enumerate*}[label={(\arabic*)}]
    \item We propose a novel loss term based on momentum-related features of motion to improve consistency between global and local motion in reconstructed human motion.
    \item We demonstrate that this term enhances the physical plausibility of motion and can be seamlessly integrated into a variety of existing models.
\end{enumerate*}

\section{Related work}
\label{sec:related}
\vspace*{-2mm}
\paragraph{Kinematics-based methods.}
Motion models are learned models that encapsulate knowledge about how humans typically move. This knowledge can be leveraged to make future predictions~\cite{Barsoum2017HPGANP3,Yuan2020DLowDL}, fill in missing data~\cite{HernandezRuiz2018HumanMP,Tevet2022HumanMD}, or constrain solutions in tasks related to human motion. For example, many optimization-based methods for pose estimation or motion generation rely on motion models to provide feedback on how plausible a predicted or generated motion is ~\cite{Kocabas2023PACEHA,Henning2022BodySLAMJC,Karunratanakul2023OptimizingDN}. 
Kinematics-based methods map motion sequences~\cite{he2022nemf,Li2021TaskGenericHH,Kocabas2019VIBEVI,petrovich21actor,petrovich22temos,zhang2023generating,Chen2022ExecutingYC} or frame-to-frame transitions~\cite{Rempe2021HuMoR3H,Henter2019MoGlow} to the latent space of generative models using only kinematic features such as joint positions, rotations, or velocities.
While effective for tasks like motion prediction in the root frame~\cite{MartnezGonzlez2021PoseT,Xu2023EqMotionEM} or reconstruction in the camera frame~\cite{Baradel2022PoseBERTAG,Goel2023HumansI4}, these models ignore the underlying dynamics of motion, often resulting in implausible global behaviors such as foot sliding or unrealistic root joint acceleration in mid-air.
Our work complements these models by introducing a simple loss term that encourages physically consistent motion through global momentum features.
\vspace*{-5mm}
\paragraph{Physics-based methods.}
Recognizing the limitation of relying on kinematics alone, many methods have increasingly made use of physical laws in both regression-based and optimization-based approaches.
Some methods use physics-related constraints to post-process generated or reconstructed motion samples, for example, by adhering more faithfully to contact events with the ground or estimating joint torques and ground reaction forces~\cite{Shimada2020PhysCap,Xie2021PhysicsbasedHM,Gartner2022DifferentiableDF,Grtner2022TrajectoryOF,Rempe2020ContactAH}. The common pipeline makes use of a physics simulator to advance the estimated kinematic pose to the next time instant given these constraints, denoising the kinematic estimation in the process~\cite{Gartner2022DifferentiableDF,Shimada2020PhysCap,Yuan2021SimPoESC,Yuan2022PhysDiffPH}.
Post-processing is challenging and might distort the generated motion enough to lose its realism.
An additional challenge is posed by the difficulty of integrating physics simulator in the pipeline, since contact events are usually discrete and, therefore, not amenable to easy inclusion in the usual machine learning model. 

To address this challenge, recent works propose to replace the physics simulator with neural components that assume the same duties, such as contact detection and force calculation~\cite{Shimada2021NeuralM3,li2022dnd,Zhang2024PhysPTPP,Zhang_2024_WACV}. They generally formulate the optimization problem with a smooth contact model to make the whole system differentiable. The estimated forces and contacts are encouraged to generate motions that match kinematics estimations, however the ground truth forces and torques are themselves generated from inverse dynamics and optimization procedure, which are sometimes unreliable due to modeling error~\cite{Yuan2020ResidualFC,Werling2024AddBiomechanicsDC}. 

Another general approach avoids relying on unobservable quantities such as forces and torques by instead examining alternative markers of motion plausibility, for example balance~\cite{Tripathi20233DHP,tripathi2024humos}, ground interaction pattern~\cite{Ma2023GraMMaRGM}, smoothness and friction~\cite{Zhang2021LearningMP}. While these methods are simpler and less computationally expensive than simulator-based counterparts, their heavy reliance on ground interaction patterns overlooks broader dynamical consistency. In contrast, our approach explicitly constrains momentum profiles, ensuring physically plausible movement across the entire body.
Centroidal dynamics have been explored in past works for motion planning~\cite{Kwon2020FastAF,Winkler2018GaitAT}, however they focus on trajectory optimization rather than directly enforcing momentum consistency in generated motions.

\section{Methodology}
\label{sec:method}

\subsection{Preliminaries}\label{sec:prelim}
\paragraph{SMPL body partition.} The center of mass of the entire body changes with different joint configurations, but each body part's center of mass remains approximately fixed relative to that body part as the body moves. 
A system's linear momentum and angular momentum are sums of the respective momenta of its parts. 
Therefore, we can calculate the momentum terms by dividing the body into parts and pre-compute some commonly needed quantities.

We partition the SMPL \cite{Loper2015SMPLAS} body mesh into $P=20$ parts. For each body part, we 
compute the volume $V_i$, the centroid $c_i^0$ and the moment of inertia $I^0_i$ of part $i$-th at the canonical pose. 
We normalize the mass by setting all subjects' total mass to $1$. Assuming a uniform distribution of mass, the mass of each part $m_i$ is then:
    $m_i = 1 \cdot \frac{V_i}{\sum_{i=1}^P V_i}$.
\vspace*{-5mm}
\paragraph{Variables definition.} Let $W$ be an inertial frame fixed to the world, its origin coincides with the root position and orientation at $t=0$. We define frame $B$ as the non-rotating frame that translates with the CoM while maintaining instantaneous alignment in orientation with the world frame $W$. All variables are in frame $W$ unless otherwise indicated. $R(t)$ is the global rotation of the body; $T(t)$ is the global translation of the body; $\theta_i(R)$ is the rotation of body part $i$-th, $c_i^W(\theta_i, T)$ and $c_i^B(\theta_i)$ are the centroid position of body part $i$-th in frame $W$ and frame $B$ respectively, $I_i(\theta_i)$ is the moment of inertia of part $i$ at $t$, where
$I_i(\theta_i) = \theta_iI^0_i\theta_i^{-1}$.
Predicted quantities are denoted with a hat (e.g. $\hat{R}$ denotes the estimated global rotation) while their ground-truth counterparts are written without it. We omit time index for simplicity.

\vspace*{-5mm}
\paragraph{Body linear momentum.}
The linear momentum of the whole human body is the vector sum of the linear momenta of all its constituent parts and is therefore: 
\abovedisplayskip=1pt
\begin{equation}\label{eq:momentum}
    \mathrm{LMo}(R, T, \theta) = \sum_{i=1}^P m_i\dot{c_i}^W.
\end{equation}
\vspace*{-5mm}
\paragraph{Body angular momentum.} Similarly, the body angular momentum is given by:
\abovedisplayskip=1pt
\begin{equation}\label{eq:wbam}
\mathrm{AMo}(R, \theta) = \sum_{i=1}^P{I_i\dot{\theta_i} + m_ic_i^B\times \dot{c_i}^B},    
\end{equation}
where the angular momentum of each body part is composed of the angular momentum about its own CoM, plus the angular momentum of its CoM about the chosen origin.

\subsection{Frequency characteristics of momentum}\label{sec:freq_char}
The quasi-periodic nature of the spin angular momentum in walking \cite{vanDien2025SimultaneousSF} suggests that the frequency decomposition of this signal may hold information about the underlying dynamics. While the class of motions we investigate is broader and may lack the same quasi-periodic structure, we still anticipate that this signal in the frequency domain exhibits certain patterns. 

Based on our analysis below, one identifiable characteristic of motion signal in frequency domain is the small magnitudes of its high-frequency components.  Angular momentum $L$ changes when there is a net external torque $\tau \neq 0$:
\begin{equation}\label{eq:torque}
\tau = \frac{dL}{dt} \therefore \Delta L(t) = L_t - L_0 = \int_0^t\tau(t) dt =  \int_{-\infty}^t\tau(t) dt. \nonumber    
\end{equation}
Therefore, let $\mathcal{F}(f(t))$ be $f(t)$ in frequency domain and $\widehat{f}(w)$ be the complex number conveying the amplitude and phase of $\mathcal{F}(f(t))$ at frequency $w$,  %\et{we can analyze the frequency characteristics of angular momentum via}:
\begin{equation}\label{eq:magnitude}
\mathcal{F}(\Delta L(t)) = \mathcal{F}\left(\int_{-\infty}^t\tau(t) dt\right) \implies \widehat{\Delta L}(w) = \frac{1}{jw} \widehat{\tau}(w). \nonumber    
\end{equation}

\noindent Since $\frac{1}{\left| w\right|}$ is a decreasing function, the magnitude of $\widehat{\Delta L}(w)$ for higher frequency $w$ is dampened compared to $\left|\widehat{\tau}(w)\right|$, which have been observed to be small \cite{Boehm2019FrequencydependentCO}. %, and is therefore expected to be negligible. 
The same analysis applies to linear momentum $p$ since $p$ only changes when there is a net external force $F \neq 0$ acting on the system, specifically $F = \frac{dp}{dt}$.

\subsection{Total momentum loss}\label{sec:total_loss}

Generated motions must not only match real kinematics but also exhibit consistent dynamics. While motion models capture kinematic patterns, they often neglect dynamics such as ground reaction forces and torques, which are hard to measure at scale. To address this, we introduce a loss term that aligns the linear and angular momentum profiles of the generated motions with the ground truth. Additionally, we constrain the power spectrum of the whole-body angular momentum to match that of the real data, following the discussion in Section \ref{sec:freq_char}.

We define $\Delta_{\mathrm{AMo}}$ and $\Delta_{\mathrm{LMo}}$ as the difference between the momentum terms of recovered motion and ground truth:
\begin{subequations}\label{eq:loss-com-def}
\begin{align}
\Delta_{\mathrm{AMo}} &= \mathrm{AMo}(\hat{R}, \hat{\theta}) - \mathrm{AMo}(R, \theta),   \\
\Delta_{\mathrm{LMo}} &= \mathrm{LMo}(\hat{R}, \hat{T}, \hat{\theta}) - \mathrm{LMo}(R, T, \theta),   
\end{align}
\end{subequations}
where $\mathrm{AMo}(R, \theta)$ is defined in Eq. \ref{eq:wbam} and $\mathrm{LMo}(R, T, \theta)$ is defined in Eq. \ref{eq:momentum}. We define the components of our loss as followed:
\begin{subequations}\label{eq:loss-def}
\begin{align}
\mathcal{L}_{\mathrm{AMo}} &= \norm{\Delta_{\mathrm{AMo}}}_2 + \norm{\Delta^\prime_{\mathrm{AMo}}}_2, \\
\mathcal{L}_{\mathrm{LMo}} &= \norm{\Delta_{\mathrm{LMo}}}_2 + \norm{\Delta^\prime_{\mathrm{LMo}}}_2, \\
\mathcal{L}_{\mathrm{S}} &= \norm{\mathcal{F}(\mathrm{AMo}(\hat{R}, \hat{\theta})) - \mathcal{F}(\mathrm{AMo}(R, \theta))}_2,
\end{align}
\end{subequations}
where $f^\prime$ denotes the time derivative of $f$ and $\mathcal{F}(f)$ is $f$ in frequency domain.
Our proposed additional loss term for training motion models is 
\begin{equation} \label{eq:total}
    \mathcal{L}_{\mathrm{TMo}} = \lambda_{\mathrm{AMo}}\mathcal{L}_{\mathrm{AMo}} 
    + \lambda_{\mathrm{LMo}}\mathcal{L}_{\mathrm{LMo}} 
 + \lambda_{\mathrm{S}}\mathcal{L}_{\mathrm{S}}.
\end{equation}
We use the discrete Fourier transform $\mathcal{F}$ and the discrete cosine transform \cite{Strang1999TheDC} for the spectrum loss $\mathcal{L}_{\mathrm{S}}$.

\section{Experiments}
\label{sec:experiment}
To evaluate the effectiveness of our loss function in increasing the consistency between global trajectory and local joint movement, we conduct experiments on the task of estimating global motion from the observations of joint motion in the root frame. This task is closely related to global motion reconstruction~\cite{Shin2023WHAMRW}, a topic that has recently gained significant attention. In this setting, we exclude any information that could be inferred from the background to more clearly isolate the impact of our loss function. 

We then demonstrate that incorporating our loss function into an existing method that leverages background information - such as camera movement - for global motion reconstruction improves motion plausibility. %without compromising global trajectory accuracy.
\vspace{-5mm}
\paragraph{Datasets.} Following GLAMR \cite{Yuan2021GLAMRGO}, for the first task, we train baseline and our variants on AMASS \cite{Mahmood2019AMASSAO}. We  %which combines several marker-based MoCap datasets for training.
evaluate on  
EMDB \cite{Kaufmann2023EMDBTE} and the Kungfu subset of Motion-X \cite{Lin2023MotionXAL}. For the second task, following WHAM \cite{Shin2023WHAMRW}, we pretrain the models on AMASS \cite{Mahmood2019AMASSAO} and finetune on 3DPW \cite{Marcard2018RecoveringA3}, Human3.6M \cite{Ionescu2014Human36MLS}, MPI-INF-3DHP \cite{Mehta2016Monocular3H} and InstaVariety \cite{Kanazawa2018Learning3H}. We evaluate on EMDB \cite{Kaufmann2023EMDBTE} and RICH \cite{Huang2022CapturingAI}. For detailed datasets description, please refer to our supplementary.
\vspace{-5mm}
\paragraph{Evaluation metrics.} There are two different aspects to the performance of a motion recovery system: accuracy, reflected by root translation error (RTE); and plausibility, measured by foot skating, acceleration and jitter. For more detailed description and formulae, please refer to our supplementary material.

\subsection{Global trajectory from local joint rotations}
\vspace{-1mm}
\paragraph{Baseline.} In this task, we evaluate the performance of the global trajectory predictor from GLAMR \cite{Yuan2021GLAMRGO} against a version of the same predictor trained with our loss function $\mathcal{L}_\mathrm{TMo}$. We also compare the performance of our CVAE based predictor against PhysPT \cite{Zhang2024PhysPTPP}, which recovers plausible global motion from an initial kinematics estimate by training a transformer model with both kinematics-based and dynamics-based  loss terms. 

\begin{figure}[ht]
\centering
\begin{minipage}[c]{0.55\linewidth}
       \includegraphics[width=\linewidth]
                   {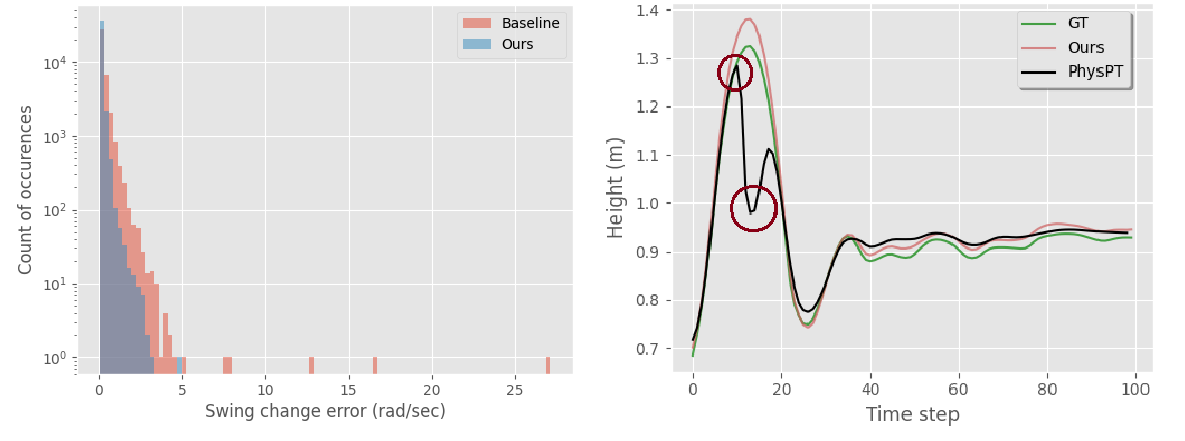}

\end{minipage}%
\hfill
\begin{minipage}[c]{0.43\linewidth}
    \caption{Left: improvement in body stability, reflected by error of the change in the swing around gravity. Right: Root joint height during a jumping sequence. PhysPT \cite{Zhang2024PhysPTPP} predicts unnatural changes of the root trajectory along the gravity direction, highlighted by red circles. %The lower circle on the right plot indicates a bounce in mid-air, which is implausible. %It is difficult for foot sliding and jitter measures to conclusively detect this kind of errors, but the deviation in high-frequency components profile of linear momentum was able to do so.
}
\label{fig:physpt-vs-ours-fig}
\end{minipage}
\end{figure}

\vspace{-5mm}
\paragraph{Quantitative results.} 
Table \ref{table:glamr} shows that our loss function improves over baselines for both plausibility linked metrics and global trajectory accuracy. 
While PhysPT \cite{Zhang2024PhysPTPP} achieves comparable jitter, the deviation of its momentum profile from the ground truth shows that there are still implausible motions generated. 
\begin{table}[ht]
\centering
\begin{minipage}[c]{0.3\linewidth}
		\caption{The predicted root trajectory from ground truth local pose. PhysPT$^\dagger$ denotes the method with the same Transformer architecture as PhysPT \cite{Zhang2024PhysPTPP}, using only position based loss and our global trajectory predictor.}
		\label{table:glamr}
\end{minipage}%
\hfill
\begin{minipage}[c]{0.65\linewidth}
\resizebox{\linewidth}{!}{%    
\begin{tabular}{l|ccc|ccc}
\toprule
\multicolumn{1}{l|}{\multirow{2}{*}{Models}}  & \multicolumn{3}{c|}{EMDB 2} & \multicolumn{3}{c}{Kungfu}                                                                              \\ %\cmidrule(l){2-6} 
\multicolumn{1}{l|}{} & \multicolumn{1}{c}{RTE$\downarrow$} & \multicolumn{1}{c}{Jitter$\downarrow$} & \multicolumn{1}{c|}{FS$\downarrow$} & \multicolumn{1}{c}{RTE$\downarrow$} & \multicolumn{1}{c}{Jitter$\downarrow$} & \multicolumn{1}{c}{FS$\downarrow$} \\ \midrule
GLAMR  & 5.19                    & 13.70                      & 7.49  & 9.89 & 40.56 & 6.86                 \\
GLAMR  + $\mathcal{L}_\mathrm{TMo}$ & \textbf{4.73}           & 7.70             & \textbf{4.91}  &  \textbf{9.16} & \textbf{34.20} & \textbf{4.81} \\
\midrule 
PhysPT  & 15.15 & 8.74 & 6.64 & 12.47 & 36.43 & 5.90 \\
PhysPT$^\dagger$  & 7.43 & 8.36 & 11.15 & 23.82 & 38.06 & 10.66 \\
PhysPT$^\dagger$  + $\mathcal{L}_\mathrm{TMo}$ & 6.58 & \textbf{5.20} & 8.22 & 13.62 & 35.06 & 7.17 \\

\end{tabular}
}
\end{minipage}
\end{table}
\vspace{-5mm}
\paragraph{Qualitative results.}
A typical improvement in body stability  with our loss is shown in Figure \ref{fig:base-vs-ours} (rightmost), even though this metric was not specifically targeted. The improved stability comes from a lack of sudden large spikes in angular momentum, which is usually observed in sequences with this type of error. See Figure \ref{fig:physpt-vs-ours-fig} (left) for the distribution change in the error of swing around gravity, indicative of body balance.

PhysPT \cite{Zhang2024PhysPTPP} focuses on human-ground contact and neglects the conservation of momentum. This can be most clearly observed in sequences involving jumping, where the foot sliding and jitter measures cannot effectively identify impossible motions such as the bounce in mid-jump shown in Figure \ref{fig:physpt-vs-ours-fig} (right).

\vspace{-5mm}
\paragraph{Perceptual study.}
To assess the increase in plausibility due to our loss, we conduct a human study on Prolific \cite{Prolific}. We select 40 sequences from AMASS and EMDB where predictions made with our version of GLAMR differ the most from the baseline. Participants are shown several videos, each consisting of two animated motions displayed side by side. They are asked to choose the motion they find more plausible. 

For each sequence, we collect 3 comparisons: ground truth vs. baseline, ours vs. baseline, and ground truth vs. ours. Each comparison is rated by 25 participants, with the same participants evaluating all comparisons for a given source sequence. The study includes two catch trials, where ground truth motions are compared with severely corrupted motions. No participants failed the catch trials.

There is a clear preference for our result compared to the baseline, with ours rated better for more than 70\% of the time, while baseline is preferred in less than 15\% of cases. Binomial test confirms that the preference for our method over the baseline is significant ($p < 0.05$). This is not solely due to the poor quality of the baseline predictions, however. We are rated as better or equal to the ground truth more than 60\% of the time. 

\subsection{Global motion recovery}
\vspace{-1mm}
\paragraph{Baseline.}
In this task, we use WHAM \cite{Shin2023WHAMRW} as the baseline.
\vspace{-5mm}
\paragraph{Quantitative results.}
SOTA comparisons for this setting is shown in Tab. \ref{table:wham}. Baseline WHAM \cite{Shin2023WHAMRW} already outperforms optimization-based method SLAHMR \cite{Ye2023DecouplingHA} and regression-based TRACE \cite{Sun2023TRACE5T}, but with our loss, it was able to reduce jitter and foot sliding error. While TRAM \cite{Wang2024TRAMGT} performs much better in global trajectory accuracy (RTE), its lack of consideration for the motion dynamic in global space leads to large errors in plausibility linked metrics.

\begin{table}[ht]
\centering
\begin{minipage}[c]{0.3\linewidth}    
		\caption{SOTA comparisons for global human motion recovery. * denotes results taken from WHAM \cite{Shin2023WHAMRW}. GLAMR$^\dagger$ uses local pose results of WHAM.}
		\label{table:wham}
\end{minipage}%
\hfill
\begin{minipage}[c]{0.65\linewidth}
\resizebox{\linewidth}{!}{%
\begin{tabular}{l|ccc|ccc}
\toprule
\multicolumn{1}{l|}{\multirow{2}{*}{Models}}  & \multicolumn{3}{c|}{EMDB 2} & \multicolumn{3}{c}{RICH}                                                                               \\ %\cmidrule(l){2-4} 
\multicolumn{1}{l|}{}  & \multicolumn{1}{c}{RTE$\downarrow$} & \multicolumn{1}{c}{Jitter$\downarrow$} & \multicolumn{1}{c|}{FS$\downarrow$} & \multicolumn{1}{c}{RTE$\downarrow$} & \multicolumn{1}{c}{Jitter$\downarrow$} & \multicolumn{1}{c}{FS$\downarrow$} \\ \midrule

GLAMR$^\dagger$ \cite{Yuan2021GLAMRGO} & 6.5 & 19.9 & 5.7 & 8.8 & 17.9 & 4.0 \\
TRACE* \cite{Sun2023TRACE5T} & 17.7 & 2987.6 & 370.7 & 610.4 & 1578.6 & 230.7\\
SLAHMR* \cite{Ye2023DecouplingHA} & 10.2 & 31.3 & 14.5 & 28.9 & 34.3 & 5.1 \\
TRAM \cite{Wang2024TRAMGT} & \textbf{1.4} & 109.1 & 20.3 & - & - & - \\
WHAM \cite{Shin2023WHAMRW}   & 4.1                    & 21.0                      & 4.4 & \textbf{4.1} & 19.7 & 3.3               \\
\midrule 
GLAMR$^\dagger$ \cite{Yuan2021GLAMRGO} + $\mathcal{L}_\mathrm{TMo}$ & 5.6 & \textbf{15.1} & \textbf{3.6} & 8.6 & \textbf{13.4} & \textbf{2.4} \\
WHAM \cite{Shin2023WHAMRW} + $\mathcal{L}_\mathrm{TMo}$ &  4.3           & 19.7             & 3.7 & 4.4 & 17.9 & 3.2
\end{tabular}
}
\end{minipage}
\end{table}

\begin{figure}[ht]
\centering
\begin{minipage}[c]{0.55\linewidth}
       \includegraphics[width=\linewidth]
                   {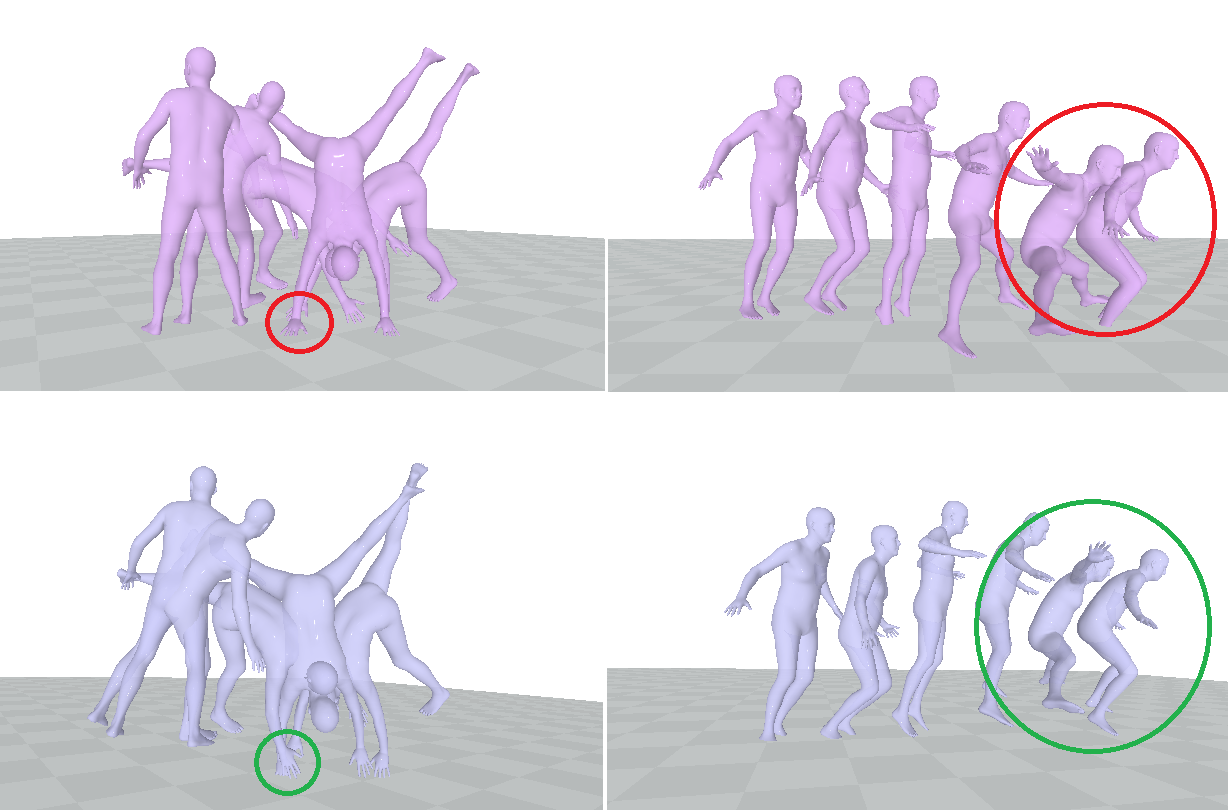}

\end{minipage}%
\hfill
\begin{minipage}[c]{0.4\linewidth}
    \caption{Qualitative examples of our improvement over WHAM \cite{Shin2023WHAMRW}. Left: during cartwheel motion, baseline did not keep the hand in correct contact with ground. Right: During a jump, baseline did not reason correctly about the height of the body's CoM.
}
\label{fig:qual-global}

\end{minipage}
\end{figure}

\vspace{-5mm}
\paragraph{Qualitative results.} Figure \ref{fig:qual-global} discusses some typical errors that our loss can address.

\vspace{-5mm}
\subsection{Ablation study}
In this section, we analyze the effect of different loss functions on the performance of the global motion inference system GLAMR \cite{Yuan2021GLAMRGO}. 
\vspace{-5mm}
\paragraph{Components} We investigate the effect of each component of $\mathcal{L}_\mathrm{TMo}$, results in table \ref{table:ablation}. Observe that $\mathcal{L}_\mathrm{TMo}$ performs well for all metrics, consistently ranked first compared to the rest. 
Regulating $\mathcal{L}_\mathrm{LMo}$ is equivalent to the control of CoM trajectory, which helps reduce root trajectory errors. We compare this loss term to directly regulating joint velocity $\mathcal{L}_\mathrm{Jv}$ (table \ref{table:other-ablation}).  While both approaches show similar performance in terms of jitter and foot sliding, $\mathcal{L}_\mathrm{LMo}$ outperforms $\mathcal{L}_\mathrm{Jv}$ in reducing RTE. 
Inspection of root rotation swing error shows that while $\mathcal{L}_\mathrm{LMo}$ can remove large outliers, $\mathcal{L}_\mathrm{AMo}$ is more effective at reducing this error overall. 
\vspace{-5mm}
\paragraph{Other related terms} Next, we compare our proposed loss to related terms to highlight its advantages. The results are reported in Table \ref{table:other-ablation}.
\begin{enumerate}[wide, labelindent=0pt]
\vspace{-2mm}
\item Root rotation can be decomposed into the twist around the gravity direction and the remaining swing. While twist can undergo drastic change quickly, swing is connected to the stability of the pose and tends to remain more stable. 
We propose the term $\mathcal{L}_{\mathrm{SW}}$ which minimizes change between the swing rotations of consecutive frames.
While reducing the sort of balance error observed in Figure \ref{fig:base-vs-ours}, compared to baseline, this loss performs worse in all reported metrics, showing that our loss does more than simply smoothing out swing rotations.
\vspace{-2mm}
\item Is it necessary to consider the shape of the body parts? To answer this question, we model the human body as a collection of point masses concentrated at the centroid of each body part and guide the model to match the angular momentum of such a system. 
\abovedisplayskip=1pt
\begin{subequations}\label{eq:transfer}
\begin{align}
        \mathrm{TF}(R, \theta) &= \sum_{i=1}^P m_ic_i^B \times \dot{c_i}^B \\
        \mathcal{L}_{\mathrm{TF}} &= \norm{\mathrm{TF}(\hat{R}, \hat{\theta}) - \mathrm{TF}(R, \theta)}_2
    \end{align}
    \end{subequations}
Compare the performance of this loss to $\mathcal{L}_\mathrm{AMo}$, we observe that $\mathcal{L}_\mathrm{AMo}$ outperforms it in all metrics, suggesting that the contribution of the angular momentum around the body parts' own CoM (and thus, the shape of the body parts) is indeed significant.

\end{enumerate}

\begin{table}[htb]
\centering
\begin{minipage}[c]{0.4\linewidth}
    		\caption{Ablation study on other related losses. }
		\label{table:other-ablation}
{\small
\begin{tabular}{c|ccc}
\toprule
      & RTE$\downarrow$ & Jitter$\downarrow$ & FS$\downarrow$ \\ \midrule
     $\mathcal{L}_{\mathrm{SW}}$ &  4.54 & 25.21 & 8.00 \\
     $\mathcal{L}_{\mathrm{TF}}$ & 4.26 & 19.24 & 6.71 \\
     $\mathcal{L}_{\mathrm{Jv}}$ & 4.60 & 19.31 & 6.55 \\
\end{tabular}
}
\end{minipage}%
\hfill
\begin{minipage}[c]{0.55\linewidth}
    		\caption{Ablation study on the loss components. 
        }
		\label{table:ablation}
{\small
\begin{tabular}{ccc|ccc}
\toprule
     $\mathcal{L}_{\mathrm{LMo}}$ & $\mathcal{L}_{\mathrm{AMo}}$ & $\mathcal{L}_{\mathrm{S}}$ & RTE$\downarrow$ & Jitter$\downarrow$ & FS$\downarrow$ \\ \midrule
     & & & 4.19 & 23.33 & 7.50 \\
     \checkmark & & & 3.90 & 19.06 & 6.51 \\
     & \checkmark & & 4.35 & 15.52 & 5.11 \\
     & & \checkmark & 4.23 & 15.53 & 5.19 \\
     \checkmark & \checkmark & & 3.85 & 13.40 & 4.10 \\
     \checkmark & \checkmark & \checkmark & 3.82 & 12.64 & 3.65
\end{tabular}
}
\end{minipage}
\end{table}

\vspace{-5mm}
\subsection{Analysis}
\vspace{-1mm}
\paragraph{Low data regime.} We investigate the effect of the amount of data available for training on the performance. We train GLAMR trajectory predictor with and without our loss on a random sample of the AMASS training set, with 20\%, 50\% and 70\% of the size of the total set. 
To facilitate comparisons between different variants, we introduce a composite measure, $m_\mathrm{AB}$ defined between two variants $A$ and $B$ as followed
\begin{equation}
    m_\mathrm{AB} = \frac{\mathrm{RTE}_A}{\mathrm{RTE}_B} \times \frac{\mathrm{Jitter}_A}{\mathrm{Jitter}_B} \times \frac{\mathrm{FS}_A}{\mathrm{FS}_B}
\end{equation}
Setting the performance of the baseline at full size as the reference, from Table \ref{table:low-data} we can see that with our loss the method can reach relatively good performance on fewer data. % check, plot?

\begin{table}[htb]
    \centering
    \begin{minipage}[c]{0.5\linewidth}
            \caption{Effect of amount of available data on performance.}
    \label{table:low-data}
    \resizebox{\columnwidth}{!}{%
    \begin{tabular}{l|cccc}
    \toprule
         Training data percentage & 0.2 & 0.5 & 0.7 & 1 \\ \midrule
         GLAMR & 3.21 & 1.49 & 1.39 & 1 \\
         GLAMR + $\mathcal{L}_{\mathrm{TMo}}$ & 0.47 & 0.28 & 0.27 & 0.25 \\
         \midrule
         Gap & 6.87 & 5.37 & 5.22 & 4.03
    \end{tabular}
    }
    \end{minipage}%
    \hfill
    \begin{minipage}[c]{0.45\linewidth}        
    \caption{The amount of extra time required by our loss compared to baseline.}
    \resizebox{\columnwidth}{!}{%
    \begin{tabular}{l|cc}
    \toprule
        Baseline & Percentage & Absolute \\
        \midrule
        GLAMR \cite{Yuan2021GLAMRGO} & 30.1\% & 6h \\
        WHAM \cite{Shin2023WHAMRW} & 11.0\% & 2h \\
        TEMOS \cite{petrovich22temos} & 2.2\% & $< 1$h \\
        PhysPT$^\dagger$ \cite{Zhang2024PhysPTPP} & 3.2\% & $<1$h\\
    \end{tabular}
    }
    \label{tab:time}

    \end{minipage}

\end{table}
\vspace{-5mm}
\paragraph{Training time.} One drawback of our loss is increased time during training. The size of the effect varies by baseline, specifically whether the components we need for calculation of the momentum terms are available, or need to be computed on the fly. The relatively large increase observed in GLAMR's trajectory predictor \cite{Yuan2021GLAMRGO} is also due to the relatively simple loss terms used by the baseline, which only involves the root trajectory.

\section{Conclusion}
\label{sec:conclusion}
We have introduced TMo, a simple yet effective loss function based on momentum features of human motion, designed to improve the physical plausibility of reconstructed motion. By encouraging consistency between global body movement and local joint dynamics through linear and angular momentum alignment, TMo addresses common artifacts such as foot sliding and unrealistic body acceleration. Applied to the global motion recovery task, TMo consistently enhances motion quality across multiple models, yielding results that are more physically realistic than baseline approaches—without compromising accuracy or requiring architectural changes.
\vspace{-5mm}
\paragraph{Limitations and future directions}
While our loss function is versatile, its application requires retraining existing models, which can be time-consuming and reliant on access to external codebases. Future work could focus on developing optimization methods that leverage a self-supervised version of our loss for easier integration. Another promising avenue is incorporating environmental context into the momentum profile optimization process.

\clearpage

\bibliography{main}
\clearpage
\clearpage
\setcounter{page}{1}
\appendix
\section{Body partition and part volume computation}
\begin{figure}[ht]
\centering
   \includegraphics[width=0.5\linewidth]
                   {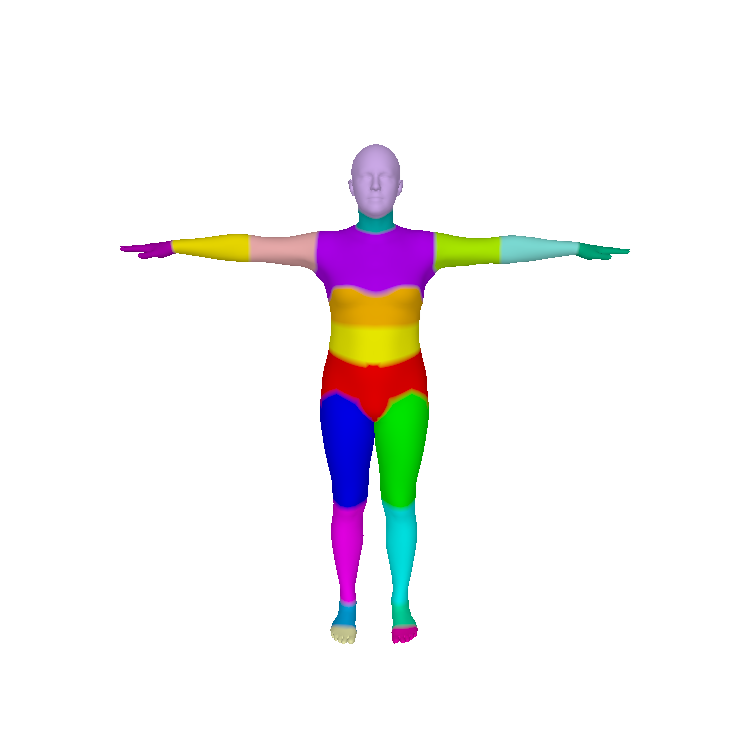}
\caption{Our partition of the SMPL body model.}
\label{fig:body-partition}
\end{figure}

\paragraph{Body partition.} The body partition we used in all experiments is based on SMPL's \cite{Loper2015SMPLAS} 24-part body partition, with the sections containing left and right shoulder merged with the upper body section. Two parts in each hand are also merged into a single entity, resulting in 20 body parts, visualized in Fig. \ref{fig:body-partition}. The boundary surface between each pair of neighboring parts is generated by adding edges between the joint connecting them and the vertices lying on the open face of each part.

While the body deforms with change in poses, the assumption that the centroid position is fixed with respect to the part's frame of reference usually holds. We sample $N=1000$ random sequences on AMASS training set and calculate the difference between the centroids position computed directly from the mesh, and computed indirectly from rotating them with respective body part's rotation. The mean error is $4.9\pm4.7$ mm, maximum $3.82$ cm.

\paragraph{Body part information.} Kallay \cite{Kallay2006ComputingTM} proposed a simple method to compute the mass properties of a solid whose boundary is defined by a mesh of triangles. We reproduced the main idea below.

Any polyhedron can be viewed as a signed sum of tetrahedra extended from the origin to its facets. 
The contribution of a triangle with vertices $(V_1, V_2, V_3)$ to the total volume of the solid is given by:
\begin{equation}
    v = \pm\frac{1}{6}\det(V_1, V_2, V_3)
\end{equation}
For an integral $\int\int\int_B f(x, y, z)dxdydz$, where $f$ is a homogeneous quadratic polynomial, the contribution of each triangle is:
\begin{equation}
    \frac{v}{20}(f(V_1) + f(V_2) + f(V_3) + f(V_1 + V_2 + V_3))
\end{equation}
Apply this to all the entries in the inertia matrix (for example, $I_{xx} = \int\int\int_B (y^2 + z^2) dm$), we recover the inertia tensor of each body part at rest pose.
\section{Evaluation metrics}
Following WHAM \cite{Shin2023WHAMRW}, we evaluate the translation error over the entire global trajectory after rigid alignment and measure Root Translation Error (RTE in \%), normalized by the actual displacement of the person. We evaluate the jitter of the motion (derivative of the acceleration) in the world coordinate system in 10$m/s^3$ and foot sliding during ground contact (in $mm$). 

\section{Datasets}
\paragraph{AMASS \cite{Mahmood2019AMASSAO}} is a large scale dataset with more than 11000 motions, unified from 15 different other marker-based MoCap datasets, with diverse motions. 
\vspace{-5mm}
\paragraph{EMDB \cite{Kaufmann2023EMDBTE}} provides global camera and body trajectories for in-the-wild videos. Following previous work, we evaluate on a subset of EMDB (25 sequences with substantial global displacement).
\vspace{-5mm}
\paragraph{Motion-X \cite{Lin2023MotionXAL}} is a large-scale 3D expressive whole-body human motion dataset from online videos and 8 existing motion datasets. We chose Kungfu subset because it has relatively difficult motion.
\vspace{-5mm}
\paragraph{3DPW \cite{Marcard2018RecoveringA3}} is an in-the-wild video dataset containing ground truth 3D pose captured by a hand-held camera and 13 body-worn inertial sensors.
\vspace{-5mm}
\paragraph{RICH \cite{Huang2022CapturingAI}} is a large-scale multi-view dataset captured in
both indoor and outdoor environment.

\section{More analysis}

\subsection{Momentum high-frequency components profile} \label{sec:momentum_hf}
The characteristic of the motion signal in frequency domain as discussed in Sec. \ref{sec:freq_char} can be used to detect implausibility in generated human motion. Motions whose momenta have high amplitude in high frequency range are less likely to be plausible. While there is a correlation between this indicator and jitter (sequences with high amplitude in high frequency range usually have high jitter, but not necessarily vice versa), an advantage this indicator has over jitter is that high jitter does not always imply an implausible motion sequence.

There are several motion plausibility measures; common ones include foot sliding, jitter, and acceleration. However, none of them reflects the dynamics principles underlying human motion, such as the fact that root movement can only occur through external forces or torques, and that momentum is conserved in the absence of such forces.

Based on 
Section \ref{sec:freq_char}, we propose a new detector of motion implausibility from the linear and angular momentum signal of the motion. Denote by $\mathcal{F}(f(t))_i$ the $i$-th component of the discrete cosine transform of the function $f(t)$. Let $H_{\mathrm{LM}}$ be the mean of the absolute magnitudes of the high-frequency components of the linear momentum:
\abovedisplayskip=1pt
\begin{equation}
    \label{eq:measure}
    H_{\mathrm{LM}} = \frac{1}{T-k_0+1}\sum_{i=k_0}^T\left|\mathcal{F}(\mathrm{LMo})_i\right|
\end{equation}
and $H_{\mathrm{AM}}$ its analog for the angular momentum.
Here, $T$ is the sequence length, $k_0$ is the threshold used to determine which component is considered part of the high-frequency range, which depends on both $T$ and the sampling frequency of the data.

We calculate $H_{\mathrm{LM}}$ and $H_{\mathrm{AM}}$ for more than $\num{40000}$ subsequences of the training set of AMASS \cite{Mahmood2019AMASSAO}, which are processed to be of the same frame rate and length, and find that both collections of these quantities are extremely leptokurtic, with kurtosis values of 460.9 and 465.5 respectively (for reference, the kurtosis value of a normal distribution is 3). These $H_{\mathrm{xM}}$ values concentrate heavily around value very close to 0, validating our hypothesis about the magnitude of high-frequency components of momentum signals in Section \ref{sec:freq_char}. 

Denote the mean of all $H_{\mathrm{LM}}$ of all sequences $\mu_{\mathrm{LM}}$ and its median absolute deviation $\sigma_{\mathrm{LM}}$, similarly for $\mu_{\mathrm{AM}}$ and $\sigma_{\mathrm{AM}}$.
We consider a sequence implausible if its $H_{\mathrm{xM}}$ falls outside the range of $\mu_{\mathrm{xM}} \pm K\sigma_{\mathrm{xM}}$. Setting $K=20$ encompasses 98.9\% of sequences in our AMASS training set. 

\begin{figure}[t!]
\centering
   \includegraphics[width=0.4\linewidth]
                   {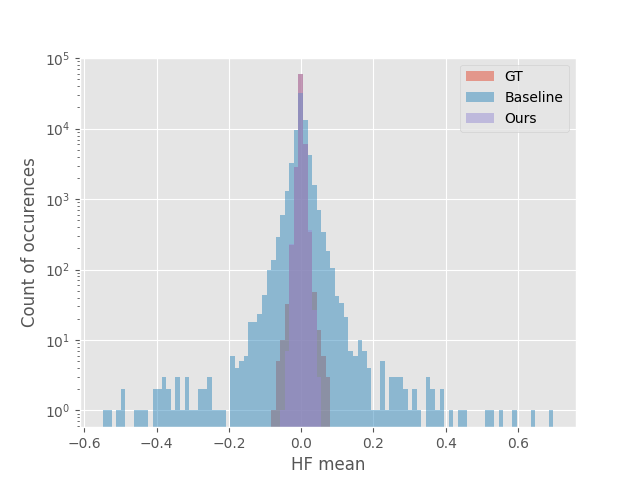}
\vskip -0.1in
\caption{The effect of our loss on angular momentum's high frequency components distribution, AMASS test set. Notice that our result follows the distribution of ground truth closely, while baseline produces angular momentum profiles with higher energy in the high frequency range. 
}
\label{fig:freq-err}
\end{figure}

\paragraph{Effect of our loss on frequency.} We illustrate the effect of our loss on the high frequency components of the angular momentum signals in Figure \ref{fig:freq-err}. While the distribution of the magnitude of our high frequency components closely follows that of the ground truth, the baseline exhibits a more spread-out pattern. These high-frequency terms with large magnitudes indicate sudden, large changes in the angular momentum, indicating possible errors.
We count the number of sequences with problematic momentum spectrum and compute their share in the population of test sequences (HF, percentage) in table \ref{table:glamr_hf}.

\begin{table}[!t]
\centering
		\caption{Root trajectory from ground truth local pose in EMDB 2. }
		\label{table:glamr_hf}
{\small
\begin{tabular}{l|cccc}
\toprule
\multicolumn{1}{l|}{\multirow{2}{*}{Models}}  & \multicolumn{4}{c}{EMDB 2}                                                                               \\ %\cmidrule(l){2-6} 
\multicolumn{1}{l|}{} & \multicolumn{1}{c}{RTE$\downarrow$} & \multicolumn{1}{c}{Jitter$\downarrow$} & \multicolumn{1}{c}{FS$\downarrow$} & \multicolumn{1}{c}{HF$\downarrow$} \\ \midrule
GLAMR  & 5.19                    & 13.70                      & 7.49  & 0.029                 \\
GLAMR  + $\mathcal{L}_\mathrm{TMo}$ & \textbf{4.73}           & \textbf{7.70}             & \textbf{4.91}  &  \textbf{0.0} \\
\midrule 
PhysPT  & 15.15 & 8.74 & \textbf{6.64} & 0.048 \\
PhysPT$^\dagger$  & 7.43 & 8.36 & 11.15 & 0.002 \\
PhysPT$^\dagger$  + $\mathcal{L}_\mathrm{TMo}$ & \textbf{6.58} & \textbf{5.20} & 8.22 & \textbf{0.0} \\

\end{tabular}
}
\end{table}

\subsection{Other ablation}

\paragraph{Noisy local motion.} We investigate the performance of our loss and baseline in the noisy setting where local joint rotations input are not accurate. Here we performs the experiment on EMDB 2 subset. We use CLIFF \cite{Li2022CLIFFCL} to recover joint rotations from images for realistic noise. Set its result as 100\% noise level, we perform SLERP between CLIFF results and ground truth data with different interpolation coefficients to simulate different levels of noise. Tab. \ref{table:noisy-data} shows that while our variant retains an advantage over baseline for all noise levels, it does not add to baseline much more robustness to noise, evidenced by the non-increasing sequence of gaps.
\begin{table}[htb]
    \centering
    \caption{Effect of noisy local data on performance, baseline on clean data as reference.}
    \vspace{0.1in}
    \label{table:noisy-data}
    \begin{tabular}{l|cccc}
    \toprule
         Noise level & 0.2 & 0.5 & 0.7 & 1 \\ \midrule
         GLAMR & 3.32 & 6.0 & 9.52 & 19.0 \\
         GLAMR + $\mathcal{L}_{\mathrm{TMo}}$ & 1.46 & 3.47 & 6.07 & 14.02 \\
         \midrule
         Gap & 3.06 & 4.12 & 2.74 & 3.13
    \end{tabular}
\end{table}

\paragraph{Directly train with plausibility measures.} Is it more effective to directly train the model to minimize measures of plausibility like jitter $\mathcal{L}_\mathrm{Jitter}$ and foot sliding $\mathcal{L}_\mathrm{FS}$? 

While $\mathcal{L}_\mathrm{Jitter}$ achieves better jitter measurement than our loss, it performs worse in both trajectory accuracy and foot sliding. Training for low foot sliding does not achieve better foot sliding performance, and is not competitive in other metrics either. Our loss improves on these metrics without explicitly guiding the model to minimize them, and does not need to sacrifice global motion accuracy.

\begin{table}[htb]
\centering
		\caption{Ablation study on related losses. }
		\label{table:other-ablation-full}
\begin{tabular}{c|ccc}
\toprule
      & RTE$\downarrow$ & Jitter$\downarrow$ & FS$\downarrow$ \\ \midrule
     $\mathcal{L}_{\mathrm{FS}}$ & 4.37 & 22.28 & 7.10 \\
     $\mathcal{L}_{\mathrm{Jitter}}$  & 5.88 & 10.47 & 8.18 \\
     $\mathcal{L}_\mathrm{TMo}$ & 3.82 & 12.64 & 3.65
\end{tabular}
\end{table}

\paragraph{Weight sensitivity.} Our loss is sensitive to weight parameters, in particular there needs to be a balance between the scale of the momentum profile loss and that of the loss controlling motion accuracy. Take the baseline performance as the reference, from Tab. \ref{table:weight-data} we can observe how the weight affects performance for different components of $\mathcal{L}_\mathrm{TMo}$.
\begin{table}[]
    \centering
    \caption{Effect of weight parameter on performance, compared to baseline.}
    \label{table:weight-data}
    {\small
    \begin{tabular}{l|ccc}
    \toprule
         Weight & 0.1 & 1.0 & 10.0 \\ \midrule
         $\mathcal{L}_\mathrm{LMo}$ & 0.66 & 0.85 & 1.56 \\
         $\mathcal{L}_\mathrm{AMo}$ & 0.71 & 0.52 & 0.47 \\
         $\mathcal{L}_\mathrm{S}$ & 0.59 & 0.47 & 0.64
    \end{tabular}
    }
\end{table}
\subsection{Perceptual study}
\begin{figure}[t!]
\centering
   \includegraphics[width=0.4\linewidth]
                   {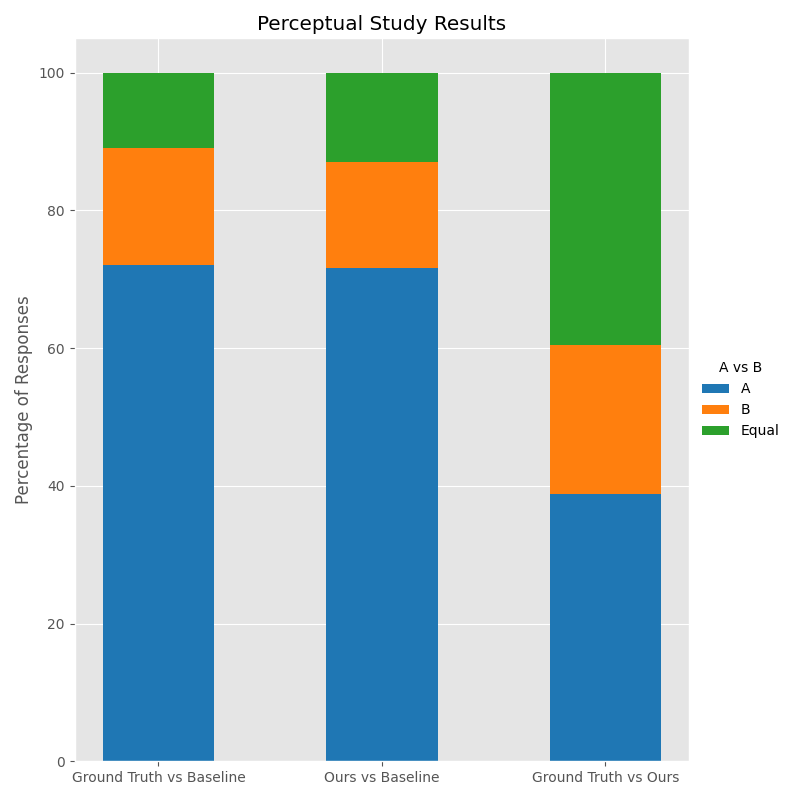}
\vskip -0.15in
\caption{Comparisons of preference rating between ground truth, baseline and our method. Each bar consists of the percentage of times each method is rated more, less or equally plausible compared to the other. Our method is judged more plausible than baseline in the majority of cases and are judged better or equally plausible to the ground truth for more than 60\% of the cases, indicating an improvement in plausibility.}
\label{fig:perceptual}

\end{figure}

We summarize the result of the perceptual study in figure \ref{fig:perceptual}.

\section{Motion generation}
We further experiment with our loss on text-to-motion generation task.

\paragraph{Baseline.}
In this task, we evaluate the performance of the motion generation model TEMOS \cite{petrovich22temos} against a version of TEMOS trained with our loss function, $\mathcal{L}_{\mathrm{TMo}}$. Additionally, we train TEMOS with a dynamics stability term recently proposed by HUMOS \cite{tripathi2024humos} to compare against TMo.

\paragraph{HUMOS stability term}
HUMOS \cite{tripathi2024humos} dynamics stability term extends the concept of pose stability proposed in IPMan \cite{Tripathi20233DHP}. Here the zero momentum point (ZMP) \cite{Vukobratovic2004ZeroMomentP} is encouraged to stay within the base of support instead of the projection of the center of mass along gravity direction in IPMan \cite{Tripathi20233DHP}. To ensure differentiability, HUMOS proposed a loss function to bring ZMP closer to the center of pressure (CoP) 
    \begin{equation}\label{eq:humos}
        \mathcal{L}_{\mathrm{HUMOS}} = \rho(\norm{\mathrm{ZMP} - \mathrm{CoP}}_2)
    \end{equation}
where $\rho$ is the robust Geman-McClure loss \cite{Geman1986}, given by:
\begin{equation}\label{eq:gm}
    \rho(x) = \frac{2x^2}{4+x^2}
\end{equation}

\paragraph{Datasets.} We train and validate TEMOS as well as other variants on KIT-ML~\cite{Plappert2016}, which combines human motion and natural language. It has $\num{6352}$ text annotations and \num{3911} motions, 900 of which are not annotated. 

\paragraph{Evaluation metrics.}
Following works in motion generation, we evaluate motion realism by FID, which computes the distribution distance between the generated and real motion on the extracted motion features using the encoder in Guo et. al \cite{Guo2022GeneratingDA}. Correspondence with text is measured by MM-Dist, which is the average Euclidean distances between each text feature and the generated motion feature from this text; and R-Precision (denoted R1), which is the top-1 accuracy of motion-to-text retrieval in the setting of choosing 1 correct text out of 32 descriptions. For diversity, we randomly sample 300 pairs of motion, computing the average Euclidean distances of these pairs. MModality computes the average Euclidean distances of 10 pairs of motions generated from the same text \cite{zhang2023generating}. 

For plausibility, the FP (floor penetration) metric measures the distance ($cm$) between the ground and the lowest body vertex below the ground. Floating measures the amount of unsupported floating by computing the distance ($cm$) between the ground and the lowest body vertex above ground. FS measures the percentage of adjacent frames where the joints in contact with the ground have an average velocity over a threshold \cite{tripathi2024humos}.

\paragraph{Quantitative results.}
Table \ref{table:temos} shows the comparison of TEMOS with our two variants on KIT-ML test set. Our variant retains and even improves the quality of motions generated with the baseline, while simultaneously improving on plausibility metrics. The variant trained with $\mathcal{L}_{\mathrm{HUMOS}}$ struggles to generate believable motions, preferring a more stable configuration throughout the sequences, which usually translate to a figure standing still. 

\begin{figure}[ht]
\centering
   \includegraphics[width=0.4\linewidth]
                   {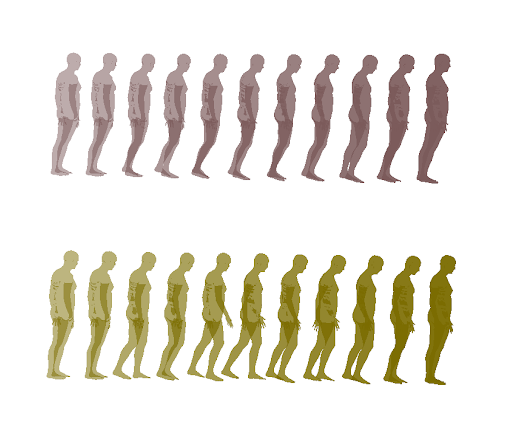}
\vskip -0.15in
\caption{Top: baseline, bottom: ours. Notice the lack of leg flexion on the baseline TEMOS motion, indicative of sliding error. }
\label{fig:temos-qual}

\end{figure}

\begin{table*}[t]
\centering
		\caption{Evaluation of TEMOS \cite{petrovich22temos} and the variants we proposed. We report the mean and 95\% confidence interval of 10 repeated evaluations. $\mathcal{L}_{\mathrm{TMo}}$ outperforms baseline on most metrics.}
		\label{table:temos}
\resizebox{\linewidth}{!}{%
\begin{tabular}{l|ccccc|ccc}
\toprule
Models & FID$\downarrow$ & R1$\uparrow$ & MM-Dist$\downarrow$ & Diversity$\uparrow$ & MModality$\uparrow$ & FS$\downarrow$ & Floating$\downarrow$ & FP$\downarrow$ \\
\midrule
TEMOS         &  $1.305^{\pm 0.039}$      &  $0.127^{\pm 0.004}$ &        $6.166^{\pm 0.020}$     & $8.042^{\pm 0.077}$                  &         $\mathbf{1.202}^{\pm 0.037}$            & 0.107 &  0.141 & 0.169                                       \\
TEMOS + $\mathcal{L}_{\mathrm{HUMOS}}$       &  $25.872^{\pm 0.152}$     &  $0.073^{\pm 0.002}$      &    $7.590^{\pm 0.017}$     &   $5.562^{\pm 0.068}$                   &    $1.404^{\pm 0.061}$               & 0.001 & 0.0 & 0.0 \\
TEMOS + $\mathcal{L}_{\mathrm{TMo}}$       &   $\mathbf{1.105}^{\pm 0.058}$  & $\mathbf{0.156}^{\pm 0.003}$           &   $\mathbf{5.820}^{\pm 0.022}$     &   $\mathbf{8.470}^{\pm 0.116}$                   &       $1.200^{\pm 0.031}$            & 0.075 & 0.099 & 0.132 \\

\end{tabular}
}
\end{table*}
\paragraph{Qualitative results.} Training TEMOS with $\mathcal{L}_\mathrm{HUMOS}$ leads to low variation in joint configurations, which results in unrealistic motions. Our loss improves on the foot sliding problem that is pervasive in TEMOS, one example is shown in Fig.\ref{fig:temos-qual}.

\end{document}